# Constructions définitoires des tables du Lexique-Grammaire

Elsa Tolone, Stavroula Voyatzi, Christian Leclère
*LIGM, Université Paris-Est*
{tolone,voyatzi,leclere}@univ-paris-est.fr

**Résumé** Les tables du Lexique-Grammaire constituent un lexique syntaxique très riche pour le français. Cette base de données linguistique n'est cependant pas directement exploitable informatiquement car elle est incomplète et manque de cohérence. Chaque table regroupe un certain nombre d'entrées jugées similaires car elles acceptent des propriétés communes qui ne sont pas codées dans les tables mais uniquement décrites dans la littérature. Notre objectif est donc de définir pour chaque table ces propriétés indispensables à l'utilisation des tables dans les diverses applications de Traitement Automatique des Langues (TAL), telle que l'analyse syntaxique.

**Mots-clés** Traitement Automatique des Langues, lexique syntaxique, Lexique-Grammaire

**Abstract** Lexicon-Grammar tables are a very rich syntactic lexicon for the French language. This linguistic database is nevertheless not directly suitable for use by computer programs, as it is incomplete and lacks consistency. Tables are defined on the basis of features which are not explicitly recorded in the lexicon. These features are only described in literature. Our aim is to define for each tables these essential properties to make them usable in various Natural Language Processing (NLP) applications, such as parsing.

## 1. Introduction

Les tables du Lexique-Grammaire[1] constituent aujourd'hui une des principales sources d'informations lexicales syntaxiques pour le français. Leur développement a été initié dès les années 1970 par (Gross, 1975), au sein du LADL puis du LIGM de l'Université Paris-Est (Boons *et al.*, 1976a; Boons *et al.*, 1976b; Guillet & Leclère, 1992). Chaque table correspond à une **classe** qui regroupe les éléments lexicaux d'une catégorie grammaticale donnée partageant certaines propriétés, que l'on appelle les **propriétés définitoires**. Elles sont en général constituées d'au moins une **construction**, dite "de base".

Les constructions de base ont subi des changements au cours des années alors que ce sont les plus importantes à définir. L'objectif de cet article est de suivre leur trace pour comprendre le sens de leur modification. Ce travail a été fait pour l'ensemble des catégories, à savoir les verbes, les noms prédicatifs, les expressions figées et les adverbes.

Nous listons tout d'abord dans la section 2, les modifications apportées aux constructions de base des classes des verbes, et plus particulièrement aux classes qui ont été dédoublées, qui n'ont jamais été publiées ou pour lesquelles on a dû spécifier la distribution de leurs arguments. Le cas des complétives et des infinitives est étudié en détail dans (Tolone, 2010). Puis, dans la section 3, nous nous penchons sur le cas de trois classes de noms avec le verbe support *avoir* afin d'illustrer les modifications apportées aux noms prédicatifs. Ensuite, dans la section 4, nous verrons le cas des expressions figées et enfin, dans la section 5, celui des adverbes en expliquant comment nous avons scindé deux ressources qui ne suivent pas les mêmes principes de représentation.

---

[1] http://infolingu.univ-mlv.fr/, Données Linguistiques > Lexique-Grammaire > Visualisation

## 2. Les classes des verbes

### 2.1 Spécification des distributions

Pour un grand nombre de classes, la construction de base est bien représentée mais il manque des informations sur la distribution spécifique d'un argument.

La construction de base de la table 32A (Apparition) décrite dans (Boons *et al.*, 1976a : 12) est *N0 V N1 apparition*, ce qui signifie que l'objet *N1* est interprété comme apparaissant ou étant créé au cours du procès. C'est le cas par exemple dans *Max a bâti une maison*. Par souci de réutilisabilité des intitulés et de simplification des distributions des arguments, nous l'avons séparée en deux propriétés définitoires : *N0 V N1* et *N1 apparition*.

La construction de base de la table 32CV (ConVersion) décrite dans (Boons *et al.*, 1976a : 14) est *N0 V N1 (E+en V-n)*, qui reconnaît par exemple *Max a roulé les papiers (E+en un mince rouleau)*. Afin d'éviter le +, ayant pour valeur *ET* ici, qui peut prêter à confusion, nous avons choisi de mettre deux constructions de base, qui sont *N0 V N1* et *N0 V N1 en V-n*. De plus, il est spécifié que tous les verbes contiennent un substantif qui dénote le résultat du procès et est interprété comme apparition après un processus de conversion (*caraméliser du sucre = le convertir/transformer en caramel*, *torsader des fils = les mettre en torsade*), c'est pourquoi nous avons également défini la propriété définitoire *N2 apparition*, le reste étant codé par les deux colonnes *V = convertir en V-n* et *V = mettre en V-n*.

La table 35S (Symétrique), décrite dans (Boons *et al.*, 1976b : 207), regroupe les verbes acceptant le double emploi *N0 V Prép N1* avec *Prép = : avec+d'avec* et *N0 et N1 V*, dont l'interprétation implique la **réciprocité**. Elle reconnaît par exemple *Max boxe avec Luc* et *Max et Luc boxent (E+ensemble+l'un avec l'autre)*. Or, les colonnes *Prép = : avec*, *Prép = : d'avec* et d'autres (telles que *dans*, *à* et *de*) sont codées dans la table, mais *Prép = : avec* étant codée + pour toutes les entrées, nous avons pu la supprimer de la table. En revanche, *Prép = : d'avec* doit figurer dans la table puisqu'elle n'est acceptée que pour certaines entrées. Nous avons donc gardé les deux constructions de base avec uniquement *Prép = : avec* définitoire.

La construction de base de la table 36DT (DaTif) décrite dans (Guillet & Leclère, 1992 : 123,237) est *N0 V N1 à N2* avec *N0 = : Nhum* et *N2 = : Nhum* (il y a une notion d'échange entre *N0* et *N2*, qui sont tous les deux humains), ce dernier pouvant se pronominaliser en *Ppv = : lui* (*lui* est pris ici comme représentant canonique de la classe des *Ppv* datives *me*, *te*, *lui*, *se*, *nous*, *vous*, *leur*) qui reconnaît par exemple *Max passe un stylo à Ida* et *Max lui passe un stylo*. En revanche, la distribution *N2 = : N-hum* étant codée dans la table, nous avons dupliqué le codage de la colonne pour la distribution *N0 = : N-hum* qui n'était pas codée. De plus, *Ppv = : lui* désigne ici la propriété définitoire mais également la propriété codée dans la table dépendante de *N2 = : N-hum*. Nous avons renommé la propriété définitoire en *Prép N2hum = Ppv = : lui*. Quant à la colonne nommée *Ppv = : lui*, nous l'avons renommée en *Prép N2-hum = Ppv = : lui* (*Ceci retire du charme à la maison* et *Ceci lui retire du charme*). De plus, comme le *N2* (qu'il soit humain ou non) peut toujours se pronominaliser en *lui*, elle contient également le même codage que *N2 = : N-hum*, soit un − lorsqu'il n'y a pas de *N2-hum*.

### 2.2. Éclatement en plusieurs classes

Comme expliqué dans (Tolone, 2010), nous avons dédoublé la table 2 car elle acceptait un argument *N1* direct pour certaines entrées et pour d'autres non. La nouvelle table 2T regroupe donc toutes les entrées transitives.

La construction de base des tables 35L (Locatif) et 35ST (STatique) décrite dans (Boons *et al.*, 1976b : 216,235) est *N0 V Loc N1*. Pour la table 35L, le *N1* est interprété comme un lieu source et/ou destination, elle reconnaît par exemple *Le bateau s'enfonce dans les flots* alors que dans la table 35ST sont regroupés les emplois **statiques** (*Le pieu sort de l'eau*). L'argument *Loc N1* de la table 35L correspond donc à un argument interprété comme une source que l'on note *N1* ou un argument interprété comme une destination noté *N2* ou les deux en même temps, ce qui pose des problèmes pour savoir à quel argument font ensuite références les propriétés codées dans la table puisque la construction de base n'en contient qu'un seul. C'est pourquoi nous avons divisé la table 35L en quatre tables (tout en séparant également les locatifs résiduels, ce qui a fait changer de table la moitié des entrées de la table 35ST) :

- 35L (Locatif) avec comme construction de base *N0 V Loc N1 source Loc N2 destination*, reconnaissant :
  *Paul a bondi du tabouret sur la table*
  *Paul a bondi du tabouret*
  *Paul a bondi sur la table*
  Ici les deux arguments sont acceptés ensemble ou chacun séparément, cela est codé dans la table par les deux constructions *N0 V Loc N1 source* et *N0 V Loc N2 destination*. Cela permet de déterminer les sources dites "dépendantes", c'est-à-dire les sources employées seulement en présence de la destination, comme dans :
  *Max chemine de chez lui vers Gap*
  **Max chemine de chez lui*
  *Max chemine vers Gap* ;
- 35LS (Locatif Source) avec comme construction de base *N0 V Loc N1 source*, c'est-à-dire ne reconnaissant pas de destination comme dans *Le train a déraillé de la voie* ;
- 35LD (Locatif Destination) ayant comme construction de base *N0 V Loc N1 destination*, c'est-à-dire ne reconnaissant pas de source comme par exemple *Le bateau s'enfonce dans les flots* ;
- 35ST (locatif STatique) avec comme construction de base *N0 V Loc N1*, rassemblant les emplois statiques des verbes de mouvement comme par exemple *Le pieu sort de l'eau* (où le pieu ne bouge pas) ou des verbes sans mouvement (*Max habite à Paris*) ;
- 35LR (Locatif Résiduel) avec comme construction de base *N0 V Loc N1*, concernant les mouvements internes à un lieu sans déplacement (*Max patauge dans la boue*).

La construction de base de la table 38LH (Locatif à corrélat Humain) décrite dans (Guillet & Leclère, 1992 : 123,202) est *N0 V N1 Loc N2,* avec *N1 = : Nhum* obligatoire (et donc *N1 = : N-hum* codée −) et dont le *N2* est interprété comme un lieu source et/ou destination. Elle reconnaît par exemple *On a viré Max de son poste*. De même que pour la table 35L, si l'on note *N2* l'argument interprété comme une source et *N3* celui interprété comme une destination, l'un, l'autre ou les deux peuvent apparaître, ce qui est contradictoire avec la construction de base. Nous avons donc divisé la table 38LH en quatre tables :

- 38LH (Locatif à corrélat Humain) avec comme construction de base *N0 V N1 Loc N2 source Loc N3 destination,* avec *N1 = : Nhum* obligatoire, qui accepte les deux arguments ensemble et reconnaît :
  *Le général a replié ses soldats du champ de bataille sur leurs lignes*
  Il y a également les deux constructions *N0 V N1 Loc N2 source* et *N0 V N1 Loc N3 destination* permettant d'accepter chacun des arguments séparément :
  *Le général a replié ses soldats du champ de bataille*
  *Le général a replié ses soldats sur leurs lignes*
  Cela permet de déterminer les sources dépendantes où seule la destination peut apparaître isolément, comme par exemple :
  *Max conduit Ida de la chambre au salon*
  **Max conduit Ida de la chambre*

*Max conduit Ida au salon* ;
- 38LHS (Locatif Source à corrélat Humain) avec comme construction de base *N0 V N1 Loc N2 source,* avec *N1 = : Nhum* obligatoire (sans destination), comme dans *On a viré Max de son poste* ;
- 38LHD (Locatif Destination à corrélat Humain) avec comme construction de base *N0 V N1 Loc N2 destination,* avec *N1 = : Nhum* obligatoire (sans source), comme par exemple *Max a engagé son fils dans la mairie* ;
- 38LHR (Locatif Résiduel à corrélat Humain) avec comme construction de base *N0 V N1 Loc N1,* avec *N1 = : Nhum* obligatoire, concernant les mouvements internes à un lieu (*Max sème Ida dans le métro*).

### 2.3. Classes de verbes non publiés

La construction de base de la table 36S (Symétrique) décrite dans l'inventaire de (Leclère, 1990) est *N0 V N1 (avec+à) N2* ou *N0 V N1 (d'avec+de) N2*, mais aussi *N0 V N1 et N2* en relation de paraphrase, qui reconnaît par exemple *Le maire a marié Paul (avec+à+et) Marie,* ou *On a dissocié Paul (d'avec+de+et) Luc*). Les colonnes *Prép2 = : à, Prép2 = : de, Prép2 = : avec* et *Prép2 = : d'avec* étant codées dans la table, nous avons noté les deux constructions de base simplement *N0 V N1 Prép N2* et *N0 V N1 et N2*.

La construction de base de la table 35RR (Résiduel double) décrite dans (Leclère, 1990) est *N0 V Prép N1 Prép N2*, qui reconnaît par exemple *Paul rivalise d'astuce avec Jean*. Cette table n'a été décrite dans aucun livre et ses intitulés sont en cours d'élaboration.

La construction de base de la table 38RR (Résiduel double) décrite dans (Leclère, 1990) est *N0 V N1 Prép N2 Prép N3*, qui reconnaît par exemple *Paul offre de l'argent à Luc pour ce travail*. Les deux prépositions sont codées dans la table, même si la plupart sont notées ~ (pas encore été codées).

### 2.4. Ajout de classe

La table 32D (Disparition) a été créée parallèlement à la table 32A (cf. 2.1) même si elle comporte peu d'entrées. L'objet *N1* est interprété comme disparaissant au cours du procès comme dans *Max a démoli la maison*. Sa construction de base est *N0 V N1,* avec *N1 disparition*. La liste des entrées codées dans cette classe est actuellement : *anéantir, démolir, détruire, fusiller, sacrifier, souffler, supprimer, volatiliser*.

## 3. Les classes des noms prédicatifs : verbe support *avoir*

Certaines classes avaient des constructions de base n'étant pas reliées aux colonnes. L'explication vient du fait que ces tables avaient des colonnes codant toutes une même construction (par exemple, *N0 avoir Det N*) avec des déterminants différents. La construction n'était pas considérée comme établie par la construction de base, mais était redéfinie à chaque fois en spécifiant en même temps la nature du déterminant. La distribution du déterminant n'était donc pas définie explicitement. Par exemple, pour la table AN09, nous avons renommé les trois colonnes *N0 avoir un N, N0 avoir un certain N* et *N0 avoir des N* respectivement par *Det = : un, Det = : un-certain* et *Det = : des*, ce qui a permis de donner un sens à la construction de base *N0 avoir Det N* définie dans (Giry-Schneider & Balibar-Mrabti, 1993 : 27).

En ce qui concerne la table AN07, sa propriété définitoire est décrite dans (Giry-Schneider & Balibar-Mrabti, 1993 : 10) par

*N0 avoir Det N (\*E+Modif)* mais également la paraphrase *N0 être de Det N (\*E+Modif)*. Elle reconnaît par exemple :
*Ce monument a une architecture (\*E+simple)*
= *Ce monument est d'une architecture (\*E+simple)*
Mais, pour les mêmes raisons que précédemment, il est préférable de séparer les informations concernant la distribution du déterminant de celles définissant la construction. C'est pourquoi nous avons défini la construction de base par *N0 avoir Det N* avec *Det = : un-Modif*. Cela nous a permis d'ajouter également la propriété définitoire *Det = : un-certain*, puisque le déterminant *un certain* est compatible avec toutes les entrées :
*Ce monument a une (\*E+certaine) architecture*
La colonne *N0 être de Det N Modif* étant codée dans la table, il n'y a pas de seconde construction de base. En effet, dans certains cas, cette construction ne s'applique pas :
*Ce pays a une vieille culture*
= *?Ce pays est d'une vieille culture*

La propriété définitoire de la table AN08 est décrite dans (Giry-Schneider & Balibar-Mrabti, 1993 : 17) par *N0 avoir Det N = il y avoir Det N Loc N0*, en précisant que *avoir = comporter+comprendre*. Elle reconnaît par exemple :
*Cette question (a+comporte+comprend) plusieurs aspects*
= *Il y a plusieurs aspects dans cette question*
La colonne *il y avoir Det N Loc N0* étant codée dans la table, nous ne l'avons pas gardée en tant que construction de base, puisqu'elle n'est pas acceptée par toutes les entrées, comme par exemple dans :
*Cette langue (a+comporte+comprend) une écriture*
= *?Il y a une écriture (dans+de) cette langue*
Nous avons donc choisi d'avoir pour la table AN08 la construction de base *N0 avoir Det N* avec les propriétés définitoires *Vsup = : comporter* et *Vsup = : comprendre*.

## 4. Les classes des expressions figées

Nous avons défini chaque construction de base en fonction de la construction morphosyntaxique interne de l'expression figée incluse dans la phrase, c'est-à-dire en ajoutant le verbe et les arguments libres qui ne sont pas inclus dans l'expression figée. Nous avons gardé la notation *N0*, *N1* et *N2* pour les arguments libres, et nous avons utilisé la notation *C0*, *C1* et *C2* pour les arguments figés. Par exemple, nous avons défini pour la table EPCPN (Gross, 1982) la construction de base *N0 être Prép1 Det1 C1 Prép2 N2*, ce qui signifie que le sujet est libre, le verbe est figé (*être*), le premier complément est figé (avec pour structure *Prép1 Det1 C1*) et le deuxième complément est libre et introduit par une préposition, comme c'est le cas pour l'entrée *être à la frontière entre*. La plupart des classes d'expressions figées sont verbales, sauf la table C0E qui contient des verbes à l'impératif (*sauve qui peut !*), des noms (*au (=à le) plaisir de vous revoir !*), des adverbes (*prochainement sur vos écrans*), etc.

La table 31I (sujet *Il*) est décrite dans (Boons *et al.*, 1976b : 263) car elle est à l'origine une classe de verbes avec comme construction de base *Il V,* qui reconnaît par exemple *Il pleut*. Nous la considérons actuellement comme une classe d'expressions figées car le sujet est figé et noté *C0*, ce qui est également le cas d'autres classes telles que par exemple EC0 dont le sujet est *ce*. De plus, comme indiqué par la distribution *N0 = : il+ça* dans (Leclère, 1990), des entrées ont été ajoutées qui n'acceptent pas le sujet *il* mais *ça*, comme par exemple *Ça dégringole*. Par ailleurs, toutes les entrées acceptant le sujet *il*, acceptent également le sujet *ça*, comme par exemple *Ça pleut*. Nous avons donc noté la construction de base *C0 V* avec *C0 = : ça*, ce qui est justifié également par le fait que la colonne *C0 = : il* est codée dans la table. Par ailleurs, une construction avec un sujet libre étant également possible pour certains verbes, la colonne *N0 V W* (*Les tomates pleuvent (E+sur la scène)*) est également codée dans la table.

## 5. Les classes des adverbes

En français, on dispose de deux ressources d'adverbes qui ne suivent pas les mêmes principes de représentation dans les tables du Lexique-Grammaire. Il s'agit, d'une part, des adverbes monolexicaux en *-ment* (Moliner, 1984; Molinier & Levrier, 2000), qui sont dérivés essentiellement des adjectifs et, d'autre part, des adverbes polylexicaux ou complexes (semi-)figés (Gross, 1986; Gross, 1990).

Pour ce qui est des adverbes (semi-)figés, la représentation dans les tables du Lexique-Grammaire est une représentation des phrases élémentaires mettant en jeu une structure de phrase simple le plus souvent à prédicat verbal intransitif dont le sujet humain (*N0 = : Nhum*) ou non-humain (*N0 = : N-hum*) est décrit et codé dans les deux premières colonnes des tables. La représentation de la combinatoire de l'adverbe avec une structure explicite de phrase élémentaire permet, d'une part, de rendre compte des relations de "portée" de l'adverbe sur un élément de la phrase :

*La réunion devra avoir lieu **au plus tard** le 15 juin* (table PAC)

Dans l'exemple ci-dessus, l'adverbe *au plus tard* porte sur l'adverbe de date *le 15 juin*. De manière générale, cet adverbe modifie obligatoirement un complément de temps.

Des contraintes de temps ou d'aspect s'observent aussi comme dans l'exemple :

*Les tablettes remplaceront les PC **dans un avenir proche*** (table PCA)
*\* Les tablettes (ont remplacé + remplacent + remplaçaient) les PC **dans un avenir proche***

Malgré leurs différences, les deux types d'adverbes sont complémentaires et sont souvent liés par des relations de paraphrase (productives et régulières) permettant de former des couples de synonymes, comme en témoignent les exemples suivants :

*pratiquement* (table ADVPS) = *en pratique* (table PC), *franchement*[2] (table ADVPS) = *à franchement parler* (table PV), *sincèrement* (table ADVMS) = *de (manière+façon) sincère* (table PCA), *politiquement* (table ADVMP) = *d'un point de vue politique* (table PCA), *malheureusement* (table ADVPAE) = *par malheur* (table PC), etc.

Toutefois, cette information significative n'apparaît pas dans les tables du Lexique-Grammaire à cause de la dispersion des adverbes concernés dans les différentes classes. Comme le signale (Gross, 1990 : 56), "seul un système de renvois explicites permettrait ces regroupements sémantiques des adverbes".

L'objectif de notre travail, entre autres, est de fournir une description complète à la fois des adverbes en *-ment* et des adverbes (semi-)figés permettant d'établir un continuum entre les deux types d'adverbes. Pour atteindre cet objectif, nous nous sommes fixé une tâche précise : l'homogénéisation de la représentation des adverbes dans les tables du Lexique-Grammaire. Les principales modifications effectuées concernent : la définition et le codage des constructions de base pour toutes les classes, l'ajout des propriétés lexicales mettant en jeu les noms et les adjectifs qui participent aux procédures transformationnelles et paraphrastiques, et l'ajout des propriétés de paraphrase, etc.

---

[2] Notons qu'il existe au moins deux emplois adverbiaux correspondant à la forme *franchement* : d'une part, adverbe de manière orienté vers le sujet appartenant à la table ADVMS, et qui peut être paraphrasé par *de manière franche* (PCA) et *avec franchise* (PC) ; et, d'autre part, adverbe de phrase disjonctif de style (ou d'énonciation) appartenant à la classe ADVPS et admettant les paraphrases *à franchement parler* (PV) et *en toute franchise* (PDETC). De manière similaire, *sincèrement* a un premier emploi en tant qu'adverbe de manière orienté vers le sujet et un deuxième emploi en tant qu'adverbe de phrase disjonctif de style (ou d'énonciation). Enfin, *pratiquement* peut être adverbe de manière verbal (ADVMV) ou adverbe de manière focalisateur (ADVMF) ou encore, adverbe de phrase disjonctif de style (ou d'énonciation) paraphrasable par *en pratique* (PC).

Les classes des adverbes monolexicaux en *-ment* de Molinier sont syntaxiquement homogènes, c'est-à-dire que chaque classe correspond à une seule classe syntaxico-sémantique des adverbes. Il a donc été facile d'attribuer les constructions de base à chaque classe. Les deux constructions *Adv, N0 V W* et *Adv, N0 ne V pas W* étaient initialement codées dans les tables des adverbes de manière et des adverbes de phrase. Nous avons donc supprimé la colonne lorsque la valeur était constante pour toutes les entrées d'une table afin de rendre compte des constructions de base.

Les classes des adverbes de manière (commençant par ADVM) ont pour construction de base *N0 V Adv W*, ce qui signifie que l'adverbe peut être placé après le verbe :
*Ce livre est en vente exclusivement sur ce site* (ADVMF, adverbes focalisateurs)
*\*Exclusivement, ce livre est en vente sur ce site*
De plus, les tables ADVMP (adverbes de point de vue), ADVMS (adverbes orientés vers le sujet) et ADVMTF (adverbes de fréquence) admettent *Adv, N0 V W* (l'adverbe peut aussi être placé en tête d'une phrase affirmative) :
*Ce livre est en vente régulièrement sur ce site* (ADVMTF)
*Régulièrement, ce livre est en vente sur ce site*
*\*Régulièrement, ce livre n'est pas en vente sur ce site*
La table ADVMP admet également *Adv, N0 ne V pas W* (l'adverbe peut aussi être placé en tête d'une phrase négative) :
*Ce concert est musicalement une réussite* (ADVMP)
*Musicalement, ce concert est une réussite*
*Musicalement, ce concert n'est pas une réussite*
Les tables des adverbes de phrase (commençant par ADVP) ont pour constructions de base *Adv, N0 V W* et *Adv, N0 ne V pas W* (l'adverbe peut être placé en tête de phrase affirmative ou négative) :
*Cinquièmement, Sunrider utilise des sous-produits animaux* (ADVPC, adverbes conjonctifs)
*Cinquièmement, Sunrider n'utilise aucun sous-produit animal*
*\*Sunrider n'utilise cinquièmement aucun sous-produit animal*
De plus, la table ADVPC admet *P1 Adv P2* (l'adverbe peut relier deux phrases).

En revanche, les classes des adverbes (semi-)figés de Gross sont syntaxiquement hétérogènes, chaque classe correspondant à une classe morphosyntaxique des adverbes polylexicaux ou complexes. Par exemple, dans la table PAC sont codés à la fois des adverbes conjonctifs (*dans un premier temps*, *en dernier lieu*), des adverbes de phrase (*à Poss0 humble avis*, *en toute première approximation*), des adverbes de temps (*en plein automne*, *depuis cent sept ans*), des adverbes de manière verbaux (*tout bêtement*, *n'importe comment*), etc. C'est pourquoi il n'est pas possible d'établir une construction de base pour une table donnée. Il faudrait en effet coder les constructions pour chaque entrée adverbiale, ce qui serait long à réaliser. Nous avons donc décidé de considérer qu'un adverbe figé pouvait se placer n'importe où dans la phrase. Les tables des adverbes figés (commençant par *P*) admettent donc les trois constructions de base 1) *N0 V Adv W*, 2) *Adv, N0 V W* et 3) *Adv, N0 ne V pas W*. De plus, nous avons défini pour chaque classe la construction morphosyntaxique interne de l'adverbe. Par exemple pour la table PCPC, l'adverbe est sous la forme *Prép1 Det1 C1 Prép2 Det2 C2*, comme c'est le cas pour l'entrée *[changer] du (=de le) jour au (=à le) lendemain*.

## 6. Conclusion

Nous avons vu que les modifications concernant les constructions de base sont diverses : certaines distributions accompagnant la construction de base étaient manquantes ; la notation + dans les constructions de base a été au maximum supprimée pour éviter une interprétation ambiguë ; certaines colonnes ont été dupliquées pour rendre compte des informations implicites de certaines classes de symétriques ; le code des arguments des constructions de base de certaines classes de locatifs a été modifié (par exemple, remplacement de *Loc N1* par *Loc N1 source* et *Loc N2 destination*), ce qui a impliqué une division des classes ; des colonnes ont été ajoutées pour rendre compte de certains cas, comme par exemple les sources dépendantes ; de nouvelles classes ont été créées ; des colonnes ont été renommées pour être reliées à la construction de base ; certaines classes acceptaient à l'origine deux constructions de base mais une seule a été retenue ; des colonnes entièrement codées + ont parfois été supprimées pour faire partie des propriétés définitoires ; certaines approximations ont été faites lorsqu'il n'existait pas de construction de base ; pour les entrées figées (expressions et adverbes figés), la structure morphosyntaxique interne de l'entrée a été définie.

Après avoir défini les constructions de base pour chaque classe dans toutes les catégories (verbes, noms prédicatifs, expressions figées et adverbes), le travail de (Tolone, 2009) a permis de construire un lexique très riche et d'envisager une utilisation de ces données lexicales dans des outils de traitement automatiques, notamment un analyseur syntaxique (Tolone & Sagot, 2009). L'utilisation d'une ressource lexicale la plus riche possible reste donc un moyen efficace pour améliorer la qualité d'un analyseur syntaxique. C'est une des motivations pour poursuivre la construction et l'amélioration des tables du Lexique-Grammaire du français et d'autres langues.

## Références